\documentclass[ draftcls,onecolumn,peerreview,12pt]{IEEEtran}

\usepackage{bm,subfigure}

\usepackage{cite}      

\usepackage{graphicx}  

\usepackage{psfrag}    

\usepackage{amsmath,amssymb}   

\usepackage{dsfont}

\newcommand{\Eq}[1]{\begin{equation}#1\end{equation}}
\newcommand{\Vc}[1]{\mbox{\boldmath$#1$}}

\newcommand{\tr}{\mbox{tr}}

\newcommand{\mR}{\mathds{R}}

\newcommand{\vx}{\Vc{x}}
\newcommand{\vphi}{\Vc{\phi}}
\newcommand{\vf}{\Vc{f}}

\newcommand{\vy}{\Vc{y}}

\newcommand{\vz}{\Vc{z}}
\newcommand{\mD}{\Vc{D}}
\newcommand{\cX}{\mathcal{X}}
\newcommand{\mX}{\Vc{X}}
\newcommand{\cY}{\mathcal{Y}}
\newcommand{\mM}{\mathcal{M}}
\newcommand{\mU}{\Vc{U}}

\newcommand{\mC}{\mathcal{C}}

\newcommand{\mZ}{\mathcal{Z}}
\newcommand{\mA}{{\bm A}}
\newcommand{\mI}{{\bm I}}

\newcommand{\mG}{{\bm G}}

\newcommand{\mH}{\Vc{H}}

\newcommand{\vu}{{\bf u}}

\newcommand{\vo}{{\bf 1}}

\newcommand{\vc}{{\bf c}}
\newcommand{\ve}{{\bf e}}
\newcommand{\vv}{{\bf v}}
\newcommand{\vs}{{\bm z}}

\newcommand{\tensor}[1]{{\bm #1}}

\newcommand{\tW}{\tensor{W}}
\newcommand{\tL}{\tensor{L}}
\newcommand{\tD}{\tensor{D}}

\newcommand{\tZ}{\tensor{Z}}
\newcommand{\tC}{\tensor{C}}
\newcommand{\diag}[1]{ {\rm diag} \{#1\}}

\begin{document}
%
\title{Classification Constrained Dimensionality Reduction}
%
%
\author{Raviv~Raich,~\IEEEmembership{Member,~IEEE,}
        Jose~A.~Costa,~\IEEEmembership{Member,~IEEE,}
        Steven~B.~Damelin,~\IEEEmembership{Senior Member,~IEEE,}
        and~Alfred~O.~Hero~III,~\IEEEmembership{Fellow,~IEEE}
\thanks{This work was partially funded by the DARPA Defense Sciences Office under Office of
Naval Research contract \#N00014-04-C-0437. Distribution Statement
A. Approved for public release; distribution is unlimited. S. B. Damelin was supported in part
by National Science Foundation grant no. NSF-DMS-0555839 and NSF-DMS-0439734 and by AFRL. }
\thanks{R. Raich is with the Oregon State University, Corvallis.  A. O Hero III is with the University of Michigan, Ann Arbor.
    J. A. Costa is with the California Institute of Technology. S. B. Damelin is with the Georgia Southern University. }}
%
%
%
\markboth{
} {Raich, Costa, and Hero: Classification Constraint
Dimensionality Reduction}

\maketitle

\begin{abstract}
Dimensionality reduction is a topic of recent interest. In this
paper, we present the classification constrained dimensionality
reduction (CCDR) algorithm to account for label information. The
algorithm can account for multiple classes as well as the
semi-supervised setting. We present an out-of-sample expressions
for both labeled and unlabeled data. For unlabeled data, we
introduce a method of embedding a new point as preprocessing to a
classifier. For labeled data, we introduce a method that improves
the embedding during the training phase using the out-of-sample
extension. We investigate classification performance using the
CCDR algorithm on hyper-spectral satellite imagery data. We
demonstrate the performance gain for both local and global
classifiers and demonstrate a $10\%$ improvement of the
$k$-nearest neighbors algorithm performance. We present a
connection between intrinsic dimension estimation and the optimal
embedding dimension obtained using the CCDR algorithm.
\end{abstract}


%
\IEEEpeerreviewmaketitle

\section{Introduction}
\label{sec:intro}

In classification theory, the main goal is to find a mapping from
an observation space ${\cal X}$ consisting of a collection of
points in some containing Euclidean space $\mR^{d}$, $d\geq 1$
into a set consisting of several different integer valued
hypotheses. In some problems, the observations from the set ${\cal
X}$ lie on a $d$-dimensional manifold $\mM$ and Whitney's theorem
tells us that provided that this manifold is smooth enough, there
exists an embedding of $\mM$ into $ \mR^{2d+1}$. This motivates
the approach taken by kernel methods in classification theory,
such as support vector machines \cite{hastie:00} for example. Our
interest is in finding an embedding of $\mM$ into a lower
dimensional Euclidean space.
\begin{figure}[h]
\centering
 \resizebox{!}{0.3\textwidth}{
 \psfrag{e1}{\large $e_1$}
 \psfrag{e2}{\large $e_2$}
 \includegraphics[width=0.5\textwidth]{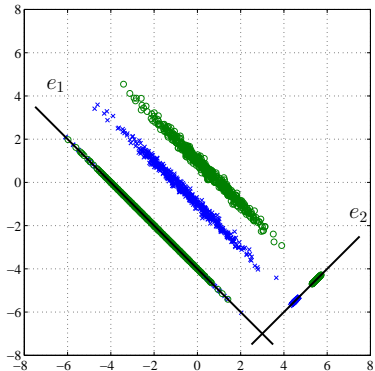}
}
 \caption{PCA of a two-classes classification problem.}
 \label{fig:pca}
\end{figure}

Dimensionality reduction of high dimensional data, was addressed
in classical methods such as principal component analysis (PCA)
\cite{Jain&Dubes:88} and multidimensional scaling (MDS)
\cite{Torgerson:52, Cox&Cox:00}. In PCA, an eigendecomposition of
the $d \times d$ empirical covariance matrix is performed and the
data points are linearly projected along the $0< m \le d$
eigenvectors with the largest eigenvalues. A problem that may
occur with PCA for classification is demonstrated in
Fig.~\ref{fig:pca}. When the information that is relevant for
classification is present only in the eigenvectors associated with
the small eigenvalues ($e_2$ in the figure), removal of such
eigenvectors may result in severe degradation in classification
performance. In MDS, the goal is to find a lower dimensional
embedding of the original data points that preserves the relative
distances between all the data points.
The two later methods suffer greatly when the manifold is nonlinear. For
example, PCA will not be able to offer dimensionality reduction
for classification of two classes lying each on one of two
concentric circles.

In \cite{Scholkopf:98:nc}, a nonlinear extension to PCA is
presented. The algorithm is based on the ``kernel trick''
\cite{aizerman:64:arc}. Data points are nonlinearly mapped into a
feature space, which in general has a higher (or even infinite)
dimension as compared with the original space and then PCA is
applied to the high dimensional data.

In the  paper of Tenenbaum \emph{et al}
\cite{Tenenbaum:00:science}, Isomap, a global dimensionality
reduction algorithm was introduced taking into account the fact
that data points may lie on a lower dimensional manifold. Unlike
MDS, geodesic distances (distances that are measured along the
manifold) are preserved by Isomap. Isomap utilizes the classical
MDS algorithm, but instead of using the matrix of Euclidean
distances, it uses a modified version of it. Each point is
connected only to points in its local neighborhood. A distance
between a point and another point outside its local neighborhood
is replaced with the sum of distances along the shortest path in
graph. This procedure modifies the squared distances matrix
replacing Euclidian with geodesic distances.

In \cite{Belkin&Niyogi:NC03}, Belkin and Niyogi present a related
Laplacian eigenmap dimensionality reduction algorithm. The
algorithm performs a minimization on the weighted sum of
squared-distances of the lower-dimensional data. Each weight
multiplying the squared-distances of two low-dimensional data
points is inversely related to distance between the corresponding
two high-dimensional data points. Therefore, small distance
between two high-dimensional data points results in small distance
between two low-dimensional data points. To preserve the geodesic
distances, the weight of the distance between two points that do
not share a local neighborhood is set to zero.

We refer the interested reader to the references below and those
cited therein for a list of some of the most commonly used
additional algorithms within the class of \emph{manifold learning}
algorithms and their different advantages relevent to our work.
Locally Linear Embedding (LLE) \cite{Roweis:00:science}, Laplacian
Eigenmaps \cite{Belkin&Niyogi:NC03}, Hessian Eigenmaps (HLLE)
\cite{Donoho:03:PNAS}, Local Space Tangent Analysis
\cite{zhang:05:siam}, Diffusion Maps \cite{coifman:06:ACHA} and
Semidefinite Embedding (SDE)\cite{weinberger:04:cvpr}.

The algorithms mentioned above, consider the problem of learning a
lower-dimensional embedding of the data. In classification, such
algorithms can be used to preprocess high-dimensional data before
performing the classification. This could potentially allow for a
lower computational complexity of the classifier. In some cases,
dimensionality reduction results increase the computational
complexity of the classifier.
In fact, support vector machines suggest the opposing strategy:
data points are projected onto a higher-dimensional space and
classified by a low computational complexity classifier. To
guarantee a low computational complexity of the classifier of the
low-dimensional data, a classification constrained dimensionality
reduction (CCDR) algorithm was introduced in
\cite{costa:05:icassp}. The CCDR algorithm is an extension of
Laplacian eigenmaps \cite{Belkin&Niyogi:NC03} and it incorporates
class label information into the cost function, reducing the
distance between points with similar label. Another algorithm that
incorporates label information is the marginal fisher analysis
(MFA) \cite{yan:07:pami}, in which a constraint on the margin
between classes is used to enforce class separation.

In \cite{costa:05:icassp} the CCDR algorithm was only studied for
two classes and its performance was illustrated for simulated
data. In \cite{raich:06:icassp}, a multi-class extension to the
problem was presented. In this paper, we introduce two additional
components that make the algorithm computationally viable. The
first is an out-of-sample extension for classification of
unlabeled test points. Similarly to the out-of-sample extension
presented in \cite{Bengio:03}, one can utilize the Nystr\"{o}m
formula for classification problems in which label information is
available.  We study the algorithm performance as its various
parameters, (e.g., dimension, label importance, and local
neighborhood),  are varied. We study the performance of CCDR as
preprocessing prior to implementation of several classification
algorithms such as $k$-nearest neighbors, linear classification,
and neural networks. We demonstrate a $10\%$ improvement over the
$k$-nearest neighbors algorithm performance benchmark for this
dataset. We address the issue of dimension estimation and its
effect on classification performance.

The organization of this paper is as follows. Section
\ref{sec:ccdr} presents the multiple-class CCDR algorithm. Section
\ref{sec:study} provides a study of the algorithm using the
Landsat dataset and Section \ref{sec:conclusion} summaries our
results.

\section{Dimensionality Reduction}
Let $\cX_n=\{\vx_1,\vx_2,\ldots,\vx_n\}$ be a set of $n$ points
constrained to lie on an $m$-dimensional submanifold $\mM
\subseteq \mR^d$.  In dimensionality reduction, our goal is to
obtain a lower-dimensional embedding $\cY_n =
\{\vy_1,\vy_2,\ldots,\vy_n\}$ (where $\vy_i \in \mR^m$ with $m<d$)
that preserves local geometry information such that processing of
the lower dimensional embedding $\cY_n$ yields comparable
performance to processing of the original data points $\cX_n$.
Alternatively, we would like learn the mapping $f: \mM \subseteq
\mR^d \to \mR^m$ that maps every data point $\vx_i$ to
$\vy_i=f(\vx_i)$ such that some geometric properties of the
high-dimensional data are preserved in the lower dimensional
embedding. The first question that comes to mind is how to select
$f$, or more specifically how to restrict the function $f$ so that
we can still achieve our goal.

\subsection{Linear dimensionality reduction}
\subsubsection{PCA}
When principal component analysis (PCA) is used for dimensionality
reduction, one considers a linear embedding of the form
\begin{eqnarray*}
\vy_i=f(\vx_i)=\mA \vx_i,
\end{eqnarray*}
where $\mA$ is $m \times d$. This embedding captures the notion of
proximity in the sense that close points in the high dimensional
space map to close points in the lower dimensional embedding,
i.e.,  $\|\vy_i-\vy_j\| = \| \mA (\vx_i - \vx_j)\| \le \|\mA\| \|
\vx_i-\vx_j\|$. Let
\[
\bar{\vx} = \frac{1}{n} \sum_{i=1}^n \vx_i
\]
and
\[
\mC_x= \frac{1}{n} \sum_{i=1}^n (\vx_i - \bar{\vx}) ( \vx_i
- \bar{\vx})^T.
\]
Similarly, let
\[
\bar{\vy} = \frac{1}{n} \sum_{i=1}^n \vy_i
\]
and
\[
\mC_y= \frac{1}{n} \sum_{i=1}^n (\vy_i - \bar{\vy}) ( \vy_i -
\bar{\vy})^T.
\]
Since $\vy_i= \mA \vx_i$, we have $\bar{\vy}= \mA \bar{\vx}$ and
$\mC_y= \mA \mC_x \mA^T$. In PCA, the goal is to find the
projection matrix $\mA$ that preserves most of the energy in the
original data by solving
\begin{eqnarray*}
\max_{\mA} \tr \{\mC_y( \mA)\} \quad \textrm{s.t.} \quad \mA\mA^T
= \mI,
\end{eqnarray*}
which is equivalent to
\begin{eqnarray}\label{eq:pca}
\max_{\mA} \tr \{\mA \mC_x \mA^T \} \quad \textrm{s.t.} \quad
\mA\mA^T = \mI.
\end{eqnarray}
The solution to (\ref{eq:pca}), is given by
$\mA=[\vu_1,\vu_2,\ldots,\vu_m]^T$, where $\vu_i$ is the
eigenvector of $\mC_x$ corresponding to its $i$th largest
eigenvalue. When the data lies on an $m$-dimensional hyperplane,
the matrix $\mC_x$ has only $m$ positive eigenvalues and the rest
are zero. Furthermore, every $\vx_i$ belongs to $  \bar{\vx}
+\textrm{span} \{\vu_1,\vu_2,\ldots,\vu_m\} \subseteq \mR^d$. In
this case, the mapping PCA finds $f(\vx)=\mA \vx$ is one-to-one
and satisfies $\|f(\vx_i)-f(\vx_j)\| = \| \mA (\vx_i - \vx_j) \| =
\| \vx_i-\vx_j \|$. Therefore, the lower embedding preserves all
the geometry information in the original dataset $\cX$. We would
like to point out that PCA can be written as
\begin{eqnarray*}
\max_{\{ \cY \}} \sum_{i=1}^n \| \vy_i-\vy_j\|^2 \quad
\textrm{s.t.} \quad \vy_i = \mA \vx_i~\textrm{and}~\mA\mA^T = \mI,
\end{eqnarray*}

\subsubsection{MDS}
Multidimensional Scaling (MDS) differs from PCA in the way the
input is provided to it. While in PCA, the original data $\cX$ is
provided, the classical MDS requires only the set of all Euclidean
pairwise distances $\{ \| \vx_i-\vx_j \|_2 \}_{i=1,j>i}^{n-1}$. As
MDS uses only pairwise distances, the solution it finds is given
up to translation and unitary transformation. Let
$\vx'_i=\vx_i-\vc$, the Euclidean distance $\|\vx'_i-\vx'_j\|$ is
the same as $\|\vx_i-\vx_j\|$. Let $\mU$ be an arbitrary unitary
matrix $\mU$ satisfying $\mU^T\mU = \mI$ and define $\vx'_i=\mU
\vx$. The distance $\|\vx'_i-\vx'_j\|$ is equal to
$\|\mU(\vx_i-\vx_j)\|$, which by the invariance of the Euclidean
norm to a unitary transformation equals to $\|\vx_i-\vx_j\|$.
Denote the pairwise squared-distance matrix by $[\mD_2]_{ij} =
\|\vx_i-\vx_j\|^2$. By the definition of Euclidean distance, the
matrix $\mD_2$ satisfies
\begin{eqnarray}\label{mds:1}
\mD_2= \vo \vphi^T + \vphi \vo^T - 2 \mX^T \mX,
\end{eqnarray}
where $\mX = [\vx_1,\vx_2,\ldots,\vx_n]$ and $ \vphi=[
\|\vx_1\|^2,\|\vx_2\|^2,\ldots, \|\vx_n\|^2]^T$. To verify
(\ref{mds:1}), one can examine the $ij$-th term of $\mD_2$ and
compare with $\| \vx_i - \vx_j \|^2$. Denote the $n \times n$
matrix $\mH = \mI - \vo \vo^T/n$. Multiplying both sides of
$\mD_2$ with $\mH$ in addition to a factor of $-\frac{1}{2}$,
yields
\[
-\frac{1}{2} \mH \mD_2 \mH = (\mX \mH)^T (\mX \mH),
\]
which is key to MDS, i.e., Cholesky decomposition of $-\frac{1}{2}
\mH \mD_2 \mH$ yields $\mX $ to within a translation and a unitary
transformation. Consider the eigendecomposition $-\frac{1}{2} \mH
\mD_2 \mH = \mU {\bm \Lambda} \mU^T$. Therefore, a rank $d$ $\mX$
can be obtained as $\mX = {\bm \Lambda}_d^{\frac{1}{2}} \mU_d^T $,
where ${\bm \Lambda}_d =
\diag{[\lambda_1,\lambda_2,\ldots,\lambda_d]}^{\frac{1}{2}}$ and
$\mU_d =  [\vu_1,\vu_2,\ldots,\vu_d]$. Note that $\mX \mH$ is a
translated version of $\mX$, in which every column $\vx_i$ is
translated to $\vx_i-\bar{\vx}$.

To use MDS for dimensionality reduction, we can consider a two
step process. First a square-distance matrix $\mD_2$ is obtained
from the high-dimensional data $\cX$. Then, MDS is applied to
$\mD_2$ to obtain a low-dimensional ($m<d$) embedding by $\mX_{m}
= {\bm \Lambda}_m^{\frac{1}{2}} \mU_m^T = \mX \mU_m \mU_m^T$. In
the absence of noise, this procedure provides an affine
transformation to the high-dimensional data and thus can be
regarded as a linear method.

\subsection{Nonlinear dimensionality reduction}

Linear maps are limited as they cannot preserve the geometry of
nonlinear manifolds.

\subsubsection{Kernel PCA}
Kernel PCA is one of the first methods in dimensionality reduction
of data on nonlinear manifolds. The method combines the
dimensionality reduction capabilities of PCA on linear manifolds
with a nonlinear embedding of data points in a higher (or even
infinite) dimensional space using ``kernel trick''
\cite{aizerman:64:arc}. In PCA, one finds the eigenvectors
satisfying: $\mC_x \vv_k = \lambda_k \vv_k$. Since $\vv_k$ can be
written as a linear combination of the $\vx_i$'s: $\vv_k = \sum_i
\alpha_{ki} (\vx_i-\bar{\vx})$, one can replace $\vv_k$ in the
eigendecomposition, simplify, and obtain: $\mX ({\bm K} {\bm
\alpha}_k - \lambda_k {\bm \alpha}_k)=0$, where ${\bm
K}_{ij}=(\vx_i-\bar{\vx})^T(\vx_j-\bar{\vx})$. Consider the
mapping ${\bm \phi}: {\cal M} \to {\cal H}$ from the manifold to a
Hilbert space. The ``kernel trick'' suggests replacing $\vx_i$
with ${\bm \phi}(\vx_i)$ and therefore rewriting the kernel as
${\bm K}_{ij}={\bm \phi}(\vx_i)^T{\bm \phi}(\vx_j)$. Further
generalization can be made by setting ${\bm
K}_{ij}=K(\vx_i,\vx_j)$ where $K(\cdot,\cdot)$ is positive
semidefinite. The resulting vectors are of the form $\vv_k =
\sum_i \alpha_{ki} {\bm \phi}(\vx_i)$ and thus implementing a
nonlinear embedding into a nonlinear manifold.

\subsubsection{ISOMAP}
 In \cite{Tenenbaum:00:science}, Tenenbaum \emph{et al}
find a nonlinear embedding that rather than preserving the
Euclidean distance between points on a manifold, preserves the
geodesic distance between points on the manifold. Similar to MDS
where a lower dimensional embedding is found to preserve the
Euclidean distances of high dimensional data, ISOMAP finds a lower
dimensional embedding that preserves the geodesic distances
between high-dimensional data points.

\subsubsection{Laplacian Eigenmaps}
Belkin and Niyogi's Laplacian eigenmaps dimensionality
reduction algorithm \cite{Belkin&Niyogi:NC03} takes a different
approach. They consider a nonlinear mapping $f$ that  minimize the
Laplacian
\begin{eqnarray}
 \arg \min_{\|f\|_{L^2(\mM)}=1} \int_{\mM} \|\nabla f \|^2.
\end{eqnarray}
Since the manifold is not available but only data point on it are,
the lower dimensional embedding is found by minimizing the graph
Laplacian given by
\begin{eqnarray}\label{laplacian}
 \sum_{i=1}^n w_{ij} \|\vy_i - \vy_j\|^2,
\end{eqnarray}
where $w_{ij}$ is the $ij$th element of the adjacency matrix which
is constructed as follows: For $k \in \mathds{N}$, a $k$-nearest
neighbors graph is constructed with the points in $\cX_n$ as the
graph vertices. Each point $\vx_i$ is connected to its $k$-nearest
neighboring points.
Note that it suffices that either $\vx_i$ is among $\vx_j$'s
$k$-nearest neighbors  or $\vx_j$ is among $\vx_i$'s $k$-nearest
neighbors for $\vx_i$ and $\vx_j$ to be connected.
For a fixed scale parameter $\epsilon > 0$, the weight associated
with the two points $\vx_i$ and $\vx_j$ satisfies
\begin{eqnarray*}
w_{ij} = \left\{ \begin{array} {l c l}

 \exp\left\{ - \| \vx_i - \vx_j \|^2 / \epsilon \right\} & \quad & \textrm {if $\vx_i$
and $\vx_j$ are connected} \\
0 & \quad & \textrm{otherwise.}
\end{array}
\right.
\end{eqnarray*}

\section{Classification Constrained Dimensionality Reduction}
\label{sec:ccdr}

\subsection{Statistical framework} To put the problem in a
classification context, we consider the following model. Let
$\cX_n=\{\vx_1,\vx_2,\ldots,\vx_n\}$ be a set of $n$ points
sampled from an $m$-dimensional submanifold $\mM \subseteq \mR^d$.
Each point $\vx_i \in \mM$ is associate with a class label $c_i
\in {\cal A}=\{0,1,2,\ldots,L\}$, where $c_i=0$ corresponds to the
case of unlabeled data. We assume that pairs $(\vx_i,c_i) \in \mM
\times {\cal A}$ are i.i.d.~drawn from a joint distribution
\begin{eqnarray}
P(\vx,c)=p_x(\vx|c) P_c(c) = P_c(c|\vx) p_x(\vx),
\end{eqnarray}
where $p_x(\vx) > 0$ and $p_x(\vx|c)>0$ (for $\vx \in \mM$) are
the marginal and the conditional probability density functions,
respectively, satisfying $\int_\mM p_x(\vx)d\vx = 1$, $\int_\mM
p_x(\vx|c) d\vx = 1 $ and $P_c(c) > 0$ and $P_c(c|\vx) > 0$ are
the a priori and a posteriori probability mass functions of the
class label, respectively, satisfying $\sum_c P_c(c) = 1$ and
$\sum_c P_c(c|\vx) = 1$. While we consider unlabeled points of the
form $(\vx_i,0)$ similar labeled points, we still make the
following distinction. Consider the following mechanism for
generating an unlabeled point. First, a class label $c \in
\{1,2,\ldots,L\}$ is generated from the labeled a priori
probability mass function $P'_c(c)=P(c|c~\textrm{is labeled}) =
P_c(c)/\sum_{c'=1}^{L} P_c(c')$. Then $\vx_i$ is generated
according to $p_x(\vx|c)$. To treat $c$ as an unobserved label, we
marginalize $P(\vx,c|c~\textrm{is labeled}) = p_x(\vx|c)P'_c(c)$
over $c$:
\begin{eqnarray}
p_x(\vx|c=0)=\sum_{q=1}^L p_x(\vx|c=q) P'_c(q) =
\frac{\sum_{q=1}^L p_x(\vx|c=q) P_c(q)}{\sum_{c'=1}^L P_c(c')}.
\end{eqnarray}
This suggests that the conditional PDF of unlabeled points
$f_x(\vx|c=0)$ is uniquely determined by the class priors and the
conditionals for labeled point. We would like to point out that
this is one of few treatments that can be offered for unlabeled
point. For example, in anomaly detection, one may want to
associate the unlabeled point with contaminated data points, which
can be represented as a density mixture of $p_x(\vx|c=0)$ and
$\gamma(\vx)$ (e.g., $\gamma(\vx)$ is uniform in $\cX$).

In classification constraint dimensionality reduction, our goal is
to obtain a lower-dimensional embedding $\cY_n =
\{\vy_1,\vy_2,\ldots,\vy_n\}$ (where $\vy_i \in \mR^m$ with $m<d$)
that preserves local geometry and that encourages clustering of
points of the same class label. Alternatively, we would like to
find a mapping ${\bm f}(\vx,c): \mM \times {\cal A} \to \mR^m$ for
which $\vy_i ={\bm f}(\vx_i,c_i)$ that is smooth and that clusters
points of the same label.

We introduce the class label indicator for data point $\vx_i$ as
$c_{ki}= I( c_i = k )$, for $k=1,2,\ldots,L$ and $i=1,2,\ldots,n$.
Note that when point $\vx_i$ is unlabeled $c_{ki}=0$ for all $k$.
Using the class indicator, we can write the number of point in
class $k$ as $n_k = \sum_{i=1}^n c_{ki}$. If all points are
labeled, then $n= \sum_{k=1}^L n_k$.

\subsection{Linear dimensionality reduction for classification}
\subsubsection{LDA}
Restricting the discussion to linear maps, one can extend PCA to
take into account label information using the multi-class
extension to Fisher's linear discriminant analysis (LDA). Instead
of maximizing the data covariance matrix, LDA maximizes the ratio
of the between-class-covariance to within-class-covariance. In
other words, we obtain a  linear transformation $\vy_i =
f(\vx_i,c_i) = \mA \vx _i$ with matrix $\mA$ that is the solution
to the following maximization:
\begin{eqnarray}\label{eq:lda}
\max_{\mA} \tr \{\mA \mC_B \mA^T \} \quad \textrm{s.t.} \quad \mA
\mC_W \mA^T  = \mI ,
\end{eqnarray}
where
\[
\mC_B = \frac{1}{n}\sum_{k=1}^L n_k (\bar{\vx}^{(k)} -
\bar{\vx})(\bar{\vx}^{(k)} - \bar{\vx})^T
\]
is the between-class-covariance matrix, $\bar{\vx}^{(k)} = \sum_i
c_{ki} \vx_i / n_k  $ is the $k$th class center, $\bar{\vx} =
\sum_i \vx_i / n  $ is the center point of the dataset,
\[
\mC_W = \frac{1}{n} \sum_{k=1}^L n_k \mC_W^{(k)},
\]
is the within-class-covariance, and
\[\mC_W^{(k)} = \frac{
\sum_{i=1}^n c_{ki} (\vx_i - \bar{\vx}^{(k)}) ( \vx_i -
\bar{\vx}^{(k)})^T }{ n_k}
\]
is within-class-$k$ covariance matrix. In Fig.~\ref{fig:pca}, LDA
selects an embedding that projects the data onto $\ve_2$ since the
maximum distance between classes is achieved along with a minimum
class variance when projecting the data onto $\ve_2$. We are
interested in exploring a strategy that maximizes class separation
in the lower dimensional embedding. \

\subsubsection{Marginal Fisher Analysis}
Recent work \cite{yan:07:pami}, presents the marginal Fisher
analysis (MFA), which is a method that minimizes the ratio between
intraclass compactness and interclass separability. In its basic
formulation MFA is a linear embedding, in which $\vy_i=\mA \vx_i$.
Another aspect of the method is that it considers two classes. The
kernel trick is used to provide a nonlinear extension to MFA. To
construct the cost function, two quantities are of interest:
intraclass compactness and interclass separability. The intraclass
compactness can be written as
\begin{eqnarray}
\sum_{i,j} w_{ij} \|\vy_i -\vy_j\|^2,
\end{eqnarray}
where $w_{ij}$ is given by
\begin{eqnarray}
w_{ij}= (\sum_k c_{ki}c_{kj}) I( \vx_i \in
N_{k_1}^+(\vx_j)~\textrm{or}~\vx_j \in N_{k_1}^+(\vx_i))
\end{eqnarray}
and $N_{k}^+(\vx)$ denote the $k$-nn neighborhood of $\vx$ within
the same class as $\vx$. Note that the term $\sum_k c_{ki}c_{kj}$
is one if $\vx_i$ and $\vx_j$ have the same label and zero
otherwise. Similarly, the interclass separability can be written
as
\begin{eqnarray}
\sum_{i,j} w_{ij} \|\vy_i -\vy_j\|^2,
\end{eqnarray}
where $w_{ij}$ is given by
\begin{eqnarray}
w_{ij}= (1-\sum_k c_{ki}c_{kj}) I( \vx_i \in
N_{k_2}^-(\vx_j)~\textrm{or}~\vx_j \in N_{k_2}^-(\vx_i))
\end{eqnarray}
and $N_{k}^-(\vx)$ denote the $k$-nn neighborhood of $\vx$ outside
the class of $\vx$.

\section{Dimensionality reduction for classification on nonlinear
manifolds} Here, we review the CCDR algorithm
\cite{costa:05:icassp} and its extension to multi-class
classification.

To cluster lower dimensional embedded points of the same label we
associate each class with a class center namely $\vz_k \in \mR^m$.
We construct the following cost function:
\Eq{
    \label{cost:function}
    J(\mZ_L,\cY_n) = \sum_{ki} c_{ki} \,
    \| \vz_k -\vy_i \|^2 + \frac{\beta}{2} \sum_{ij} w_{ij} \, \| \vy_i -\vy_j \|^2  ,
}
where $\mZ_L = \{\vz_1,\ldots,\vz_L\}$ and $\beta \geq 0$ is a
regularization parameter. We consider two terms on the RHS of
(\ref{cost:function}). The first term corresponds to the
concentration of points of the same label around their respective
class center. The second term is as in (\ref{laplacian}) or as in
Laplacian Eigenmaps \cite{Belkin&Niyogi:NC03} and controls the
smoothness of the embedding over the manifold. Large values of
$\beta$ produce an embedding that ignores class labels and small
values of $\beta$ produce an embedding that ignores the manifold
structure. Training data points will tend to collapse into the
class centers, allowing many classifiers to produce perfect
classification on the training data without being able to control
the generalization error (i.e., classification error of the
unlabeled data). Our goal is to find $\mZ_L$ and $\cY_n$ that
minimize the cost function in (\ref{cost:function}).

Let $\tC$ be the $L \times n$ class membership matrix with
$c_{ki}$ as its  $ki$-th element, $\tZ = [\vz_1 , \ldots , \vz_L ,
\vy_1, \ldots ,\vy_n]$, and $\tensor{0}$ be the $L \times L$ all
zeroes matrix and
\[
    \mG = \left[
    \begin{array}{cc}
        \tensor{0} & \tC \\
        \tC^T & \beta \tW
    \end{array}
    \right] \ .
\]
Minimization over $\tZ$ of the cost function in
(\ref{cost:function}) can be expressed as
\Eq{ \label{E:Optimization_Extended}
    \min_{\footnotesize{\begin{array}{c} \tZ \tD  \Vc{1} = \Vc{0} \\
    \tZ  \tD  \tZ^T = \tensor{I} \end{array} }} \mbox{tr} \left( \tZ  \tL  \tZ^T \right) \ ,
}
where $\tD=\diag{\mG \Vc{1}}$ and $\tL=\tD-\mG$. To prevent the
lower-dimensional points and the class centers from collapsing
into a single point at the origin, the regularization $ \tZ \tD
\tZ^T = \tensor{I} $ is introduced. The second constraint $\tZ \tD
\Vc{1} = \Vc{0}$ is constructed to prevent a degenerate solution,
e.g., $\vz_1=\ldots=\vz_L=\vy_1=\ldots=\vy_n$. This solution may
occur since $\Vc{1}$ is in the null-space of the Laplacian $\tL$
operator, i.e., $\tL\Vc{1}=\Vc{0}$. The solution to
(\ref{E:Optimization_Extended}) can be expressed in term of the
following generalized eigendecomposition
\begin{eqnarray}\label{eq:eig:1}
\tL^{(n)}  \vu_k^{(n)} =  \lambda_k^{(n)} \tD^{(n)} \vu_k^{(n)},
\end{eqnarray}
where $\lambda_k^{(n)}$ is the $k$th eigenvalue and $\vu_k^{(n)}$
is its corresponding eigenvector. Note that we include $^{(n)}$ to
emphasize the dependence on the $n$ data points. Without loss of
generality we assume $\lambda_1 \le \lambda_2 \le \ldots \le
\lambda_{n+L}$. Specifically, matrix $\tZ$ is given by $[\vu_2,
\vu_3, \ldots, \vu_{m+1}]^T$, where the first $L$ columns
correspond to the coordinates of the class centers, i.e.,
$\vz_k=\tZ \ve_k$, and the following $n$ columns determine the
embedding of the $n$ data points, i.e., $\vy_t=\tZ \ve_{L+t}$. We
use $\ve_i$ to denote the canonical vector such that $[\ve_i]_{s}=
1$ for element $s=i$ and zero otherwise.

\subsection{Classification and computational complexity}

In classification, the goal is to find a classifier $a_x(\vx):\mM
\to {\cal A}$ based on the training data that minimizes the
generalization error:
\begin{eqnarray}\label{general}
\hat{a} = \arg \min_{a \in {\cal F}} E[ I(a (\vx) \neq a)],
\end{eqnarray}
where the expectation is taken w.r.t.~the pair $(\vx, a)$. Since
only samples from the joint distribution of $\vx$ and $a$ are
available, we replace the expectation with a sample average
w.r.t.~the training data $\frac{1}{n} \sum_{i=1}^n I(a(\vx_i) \neq
a_i)$. During the minimization, we search over a set of
classifiers $a_x(\vx):\mM \subseteq \mR^d \to {\cal A}$, which is
defined over a domain in $\mR^d$. In our framework, we suggest
replacing a classifier $a_x(\vx):\mM \subseteq \mR^d \to {\cal A}$
with dimensionality reduction via CCDR $f(\vx): \mM \subseteq
\mR^d \to \mR^m$ followed by a classifier on the lower-dimensional
space $a_y({\bm y}): \mR^m \to {\cal A}$, i.e., $a_x = a_y \circ
f$.  The first advantage is that the search space for the
minimization in (\ref{general}) defined over a $d$-dimensional
space can be reduced to an $m$-dimensional space. This results in
significant savings in computational complexity if the complexity
associated with the process of obtaining $f$ can be made low. In
general, the classifier set ${\cal F}$ has to be rich enough to
attain a lower generalization error. The other advantage of our
method lies in the fact that CCDR is designed to cluster points of
the same label thus allowing for a linear classifier or other low
complexity classifiers. Therefore, further reduction in the size
of class ${\cal F}$ can be achieved in addition to the reduction
due to a lower-dimensional domain. To classify a new data point,
one has to apply CCDR to a new data point. If it is done brute
force, the point is added to the set of training points with no
label a new matrix $W'$ is formed and an eigendecomposition is
carried out.

When performing CCDR, each of the $n(n-1)/2$ terms of the form $\{
\| \vx_i - \vx_j \|^2 \}$ requires one summation and $d$
multiplications leading to computational complexity of the order
$O(dn^2)$. Construction of a $K$-nearest neighbors graph requires
$O(kn)$ comparisons per point and therefore a total of $O(kn^2)$.
The total number of operations involved in constructing the graph
is therefore $O((k+d)n^2)$. Next, an eigendecomposition is applied
to $W'$, which is an $(L+n) \times (L+n)$ matrix. The associated
computation complexity is $O(n^3)$. Therefore, the overall
computational complexity of CCDR is $O(n^3)$. This holds for both
training and classification as explained earlier. We are
interested in reducing computational complexity in training the
classifier and in classification. For that purpose, we consider an
out-of-sample extension of CCDR.

\section{Out-of-Sample Extension}

We start by rearranging the generalized eigendecomposition of the
Laplacian in (\ref{eq:eig:1}) as
\begin{eqnarray}\label{eq:eig:2}
\mG^{(n)}  \vu_l^{(n)} =  (1-\lambda_l^{(n)}) \tD^{(n)}
\vu_l^{(n)},
\end{eqnarray}
and recall that $\vu_l^{(n)}=[\vs_1(l),\vs_1(l),\ldots,\vs_1(l),
\vy_1(l), \vy_2(l), \ldots, \vy_n(l)]^T$. Since we consider an
$m$-dimensional embedding, we are only interested in eigenvectors
$\vu_2,\ldots,\vu_{m+1}$.
 The $L+i$ equation (row) for $i=1,2,\ldots,n$ in the eigendecomposition in
(\ref{eq:eig:2}) can be written as
\begin{eqnarray}\label{eq:y:sample}
\vy_i^{(n)}(l)=\frac{1}{1- \lambda_l^{(n)}} \frac{ \sum_k
c_{ki}\vs_k^{(n)}(l)+ \beta \sum_j K(\vx_i,\vx_j) \vy_j^{(n)}(l)}
{ \sum_k c_{ki}+ \beta \sum_j K(\vx_i,\vx_j)}.
\end{eqnarray}
Similarly, the $k$th  equation (row) of (\ref{eq:eig:2}) for
$k=1,2,\ldots,L$ is given by
\begin{eqnarray}\label{eq:z:sample}
\vs_k^{(n)}(l)= \frac{\sum_i c_{ki}
\vy_i^{(n)}(l)}{(1-\lambda_l^{(n)}) n_k}.
\end{eqnarray}
Our interest is in finding a mapping $\vf(\vx,c)$ that in addition
to mapping every $\vx_i$ to $\vy_i$, can perform an out-of-sample
extension, i.e., is well-defined outside the set $\cX$. We
consider the following out-of-sample extension expression
\begin{eqnarray}\label{eq:y:sample:oos}
\vf_l^{(n)}(\vx,c)=\frac{1}{1- \lambda_l^{(n)}} \frac{ I(c\neq 0)
\vs_c^{(n)}(l)+ \beta \sum_j K(\vx,\vx_j) \vy_j^{(n)}(l)} {
I(c\neq 0)+ \beta \sum_j K(\vx,\vx_j)},
\end{eqnarray}
where $\vs^{(n)}$ is the same as in (\ref{eq:z:sample}). This
formula can be explain as follows. First, the lower dimensional
embedding $\vy^{(n)}_1,\ldots, \vy_n^{(n)}$ and the class centers
$\vz_1^{(n)},\ldots,\vz_L^{(n)}$ are obtained through an the
eigendecomposition in (\ref{eq:eig:2}). Then, the embedding
outside the sample set $\cX$ is calculated via
(\ref{eq:y:sample:oos}). By comparison of $\vf_l^{(n)}(\vx_i,c_i)$
evaluated through (\ref{eq:y:sample:oos}) with
(\ref{eq:y:sample}), we have $\vf_l^{(n)}(\vx_i,c_i) =
\vy_i^{(n)}(l)$. This suggests that the out-of-sample extension
coincides with the solution, we already have for the mapping at
the the data points $\cX$. Moreover, using this result one can
replace all $\vy_i^{(n)}$ with $\vf_l^{(n)}(\vx_i,c_i)$ in
(\ref{eq:y:sample:oos}) and obtain the following generalization of
the eigendecomposition in (\ref{eq:eig:2}):
\begin{eqnarray}\label{eq:f:eig}
\vf_l^{(n)}(\vx,c)=\frac{1}{1- \lambda_l^{(n)}} \frac{ I(c\neq 0)
\vs_c^{(n)}(l)+ \beta \sum_j K(\vx,\vx_j) \vf_l^{(n)}(\vx_j,c_j)
}{ I(c\neq 0)+ \beta \sum_j K(\vx,\vx_j)},
\end{eqnarray}
and
\begin{eqnarray}\label{eq:z:f:eig}
\vs_k^{(n)}(l)= \frac{\sum_i c_{ki}
\vf_l^{(n)}(\vx_i,c_i)}{(1-\lambda_l^{(n)}) n_k}.
\end{eqnarray}
In \cite{Bengio:04}, it is propose that if the out-of-sample
solution to the eigendecomposition problem associated with kernel
PCA converge, it is given by the solution to the asymptotic
equivalent of the eigendecomposition. Using similar machinery, we
can provide a similar result suggesting that if
$\vf_l^{(n)}(\vx,c) \to \vf_l^{(\infty)}(\vx,c)$ as $n \to
\infty$, then the asymptotic equivalents to (\ref{eq:f:eig}) and
(\ref{eq:z:f:eig}) should provide the solution to the limit of
$\vf_l^{(n)}(\vx,c)$. The asymptotic analogues to
(\ref{eq:y:sample}) and (\ref{eq:z:sample}) are described in the
following. The mapping for labeled data $f_l(\vx,c): \mM \times
{\cal A} \to \mR$ for $c=0,1,2,\ldots,L$ equivalent to equation
(\ref{eq:y:sample}) is
\begin{eqnarray}\label{map:f}
f_l(\vx,c)=\frac{1}{1-\lambda_l} \frac{  I(c \neq 0) \vs_c(l)+
\beta' \sum_{c'=0}^L \int_{\mM} K(\vx,\vx') f_l(\vx',c')
P(\vx',c') d\vx'} { I(c \neq 0)  + \beta' \int_{\mM} K(\vx,\vx')
p(\vx') d\vx'}
\end{eqnarray}
where $\vs_c(l)$ for $c=1,2,\ldots,L$ is equivalent to
(\ref{eq:z:sample})
\begin{eqnarray}
\vs_c(l)= \frac{\int_{\mM} f_l(\vx,c) p(\vx|c) d \vx }
{1-\lambda_l},
\end{eqnarray}
and $\beta'=\beta n$. Since we are interested in an
$m$-dimensional embedding, we consider only $l=1,2,\ldots,m$,
i.e., the eigenvectors that correspond to the $m$ smallest
eigenvalues. To guarantee that the relevant eigenvectors are
unique (up to a multiplicative constant), we require
$\lambda_1<\lambda_2< \cdots < \lambda_{m+1} \le \lambda_{m+2} \le
\ldots \lambda_{n}$.

The out-of-sample extension given by (\ref{eq:y:sample:oos}), can
be useful in a couple of scenario. The first, is in classification
of new unlabeled samples. We assume that $\{\vy_j\}_{j=1}^n$,
$\{\vs_k\}_{k=1}^L$, and $\{ \lambda_l \}_{l=1}^m$ are already
obtained based on labeled (or partially labeled) training data and
we would like to embed a new unlabeled data point. We consider
using (\ref{eq:y:sample:oos}) with $c=0$, i.e., we can use
$\vf(\vx,0)$ to map a new sample $\vx$ to $\mR^m$. The obvious
immediate advantage is the savings in computational complexity as
we avoid performing addition eigendecomposition that includes the
new point.

The second scenario involves the out-of-sample extension for
labeled data. The goal here is not to classify the data since the
label is already available. Instead, we are interested in the
training phase in the case of large $n$ for which the
eigendecomposition is infeasible. In this case, a large amount of
labeled training data is available but due to the heavy
computational complexity associated with the eigendecomposition in
(\ref{eq:eig:1}) (or by (\ref{eq:eig:2})), the data cannot be
processed. In this case, we are interested in developing a
resampling method, which integrates $\vf^{(n)}_l(\vx, c)$ obtained
for different subsamples of the complete data set. 

\subsection{Classification Algorithms}
We consider three widespread algorithms: $k$-nearest neighbors,
linear classification, and neural networks. A standard
implementation of $k$-nearest neighbors was used, see \cite[p.
415]{hastie:00}. The linear classifier we implemented is given by
\begin{eqnarray*}
\hat{c}({\vy}) & = &
 \arg \max_{c \in \{ {\cal A}_1,\ldots {\cal A}_L\}}
 \vy^T {\bm \alpha}^{(c)} + \alpha_0^{(c)}\\
 \bigl[
 {\bm \alpha}^{({\cal A}_k)},
 \alpha_0^{({\cal A}_k)}
 \bigr]
 & = & \arg
\min_{[{\bm \alpha} , \alpha_0]} \sum_{i=1}^n (\vy_i^T {\bm
\alpha} + \alpha_0 - c_{ki})^2,
\end{eqnarray*}
for $k=1,\ldots,L$. The neural network we implemented is a
three-layer neural network with $d$ elements in the input layer,
$2d$ elements in the hidden layer, and $6$ elements in the output
layer (one for each class). Here $d$ was selected using the common
PCA procedure, as the smallest dimension that explains $99.9\%$ of
the energy of the data. A gradient method was used to train the
network coefficients with 2000 iterations. The neural net is
significantly more computationally burdensome than either linear
or $k$-nearest neighbors classifications algorithms.

\subsection{Data Description}
In this section, we examine the performance of the classification
algorithms on the benchmark label classification problem provided
by the Landsat MSS satellite imagery database \cite{sat:data}.
Each sample point consists of the intensity values of one pixel
and its 8 neighboring pixels in 4 different spectral bands. The
training data consists of 4435 36-dimensional points of which,
1072 are labeled as 1) red soil, 479 as 2) cotton crop, 961 as 3)
grey soil, 415 as 4) damp grey soil, 470 are labeled as 5) soil
with vegetation stubble, and 1038 are labeled as 6) very damp grey
soil. The test data consists of 2000 36-dimensional points of
which, 461 are labeled as 1) red soil, 224 as 2) cotton crop, 397
as 3) grey soil, 211 as 4) damp grey soil, 237 are labeled as 5)
soil with vegetation stubble, and 470 are labeled as 6) very damp
grey soil. In the following, each classifier is trained on the
training data and its classification is evaluated based on the
entire sample test data. In Table~\ref{table:performance}, we
present ``best case'' performance of neural networks, linear
classifier, and $k$-nearest neighbors in three cases: no
dimensionality reduction, dimensionality reduction via PCA, and
dimensionality reduction via CCDR. The table presents the minimum
probability of error achieved by varying the tuning parameters of
the classifiers. The benefit of using CCDR is obvious and we are
prompted to further evaluate the performance gains attained using
CCDR.
\begin{table}[htb!]
\begin{tabular}{l|r|r|r}
 &         Neural Net.  &  Lin.  & $k$-nearest neigh. \\ \hline
 No dim. reduc.   &   83 \%  &   22.7 \% & 9.65 \% \\
 PCA              & 9.75 \%  &     23 \% & 9.35 \% \\
 CCDR             & 8.95 \%  &   8.95 \% &  8.1 \%
\end{tabular}
\caption{Classification error probability}
 \label{table:performance}
\end{table}

\subsection{Regularization Parameter $\beta$}\label{sec:beta}
\begin{figure}[htb!]
 \centering
 \resizebox{!}{0.33\textwidth}{
 \psfrag{beta}{\large $\beta$}
 \psfrag{P(error)}{P(error)}
 \includegraphics[width=0.5\textwidth]{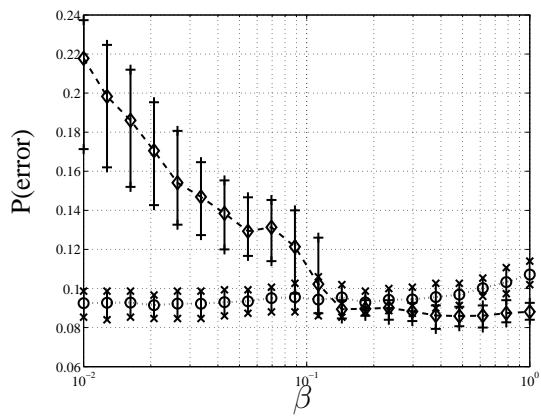}
 }
\caption{Probability of incorrect classification vs. $\beta$
 for a linear classifier (dotted line $\circ$) and
 for the $k$-nearest neighbors algorithm (dashed line $\diamond$) preprocessed by CCDR.
 $80\%$ confidence intervals are presented as $\times$ for the linear classifier and as $+$
 for the $k$-nearest neighbors algorithm.}
 \label{fig:vs:beta}
\end{figure}

As mentioned earlier, the CCDR regularization parameter $\beta$
controls the contribution of the label information versus the
contribution of the geometry described by the sample. We apply
CCDR to the 36-dimensional data to create a 14-dimensional
embedding by varying $\beta$ over a range of values. For
justification of our choice of $d=14$ dimensions see Section
\ref{sec:dim}. In the process of computing the weights $w_{ij}$
for the algorithm, we use $k$-nearest neighbors with $k=4$ to
determine the local neighborhood. Fig.~\ref{fig:vs:beta} shows the
classification error probability (dashed lines) for the linear
classifier vs. $\beta$ after preprocessing the data using CCDR
with $k=4$ and dimension 14.
We observe that for a large range of $\beta$ the average
classification error probability is greater than $0.09$ but
smaller than $0.095$. This performance competes with the
performance of $k$-nearest neighbors applied to the
high-dimensional data, which is presented in \cite{hastie:00} as
the leading classifier for this benchmark problem. Another
observation is that for small values of $\beta$ (i.e.,
$\beta<0.1$) the probability of error is constant. For such small
value of $\beta$, classes in the lower-dimensional embedding are
well-separated and are well-concentrated around the class centers.
Therefore, the linear classifier yields perfect classification on
the training set and fairly low constant probability of error on
the test data is attained for low value of $\beta$. When $\beta$
is increased, we notice an increase in the classification error
probability. This is due to the fact that the training data become
non separable by any linear classifier as $\beta$ increases.

We perform a similar study of classification performance for
$k$-nearest neighbors. In Fig.~\ref{fig:vs:beta}, classification
probability error is plotted (dotted lines) vs. $\beta$. Here, we
observed that an average error probability of $0.086$ can be
achieved for $\beta\approx 0.5$. Therefore, $k$-nearest neighbors
preceded by CCDR outperforms the straightforward $k$-nearest
neighbors algorithm. We also observe that when $\beta$ is
decreased the probability of error is increased. This can be
explained as due to the ability of $k$-nearest neighbors to
utilize local information, i.e., local geometry. This information
is discarded when $\beta$ is decreased.

We conclude that CCDR can generate lower-dimensional data that is
useful for global classifiers, such as the linear classifier, by
using a small value of $\beta$, and also for local classifiers,
such as $k$-nearest neighbors, by using a larger value $\beta$ and
thus preserving local geometry information.
\subsection{Dimension Parameter}\label{sec:dim}
\begin{figure}[htb!]
 \centering
 \resizebox{!}{0.33\textwidth}{
 \psfrag{dim.}{dim.}
 \psfrag{P(error)}{P(error)}
 \includegraphics[width=0.5\textwidth]{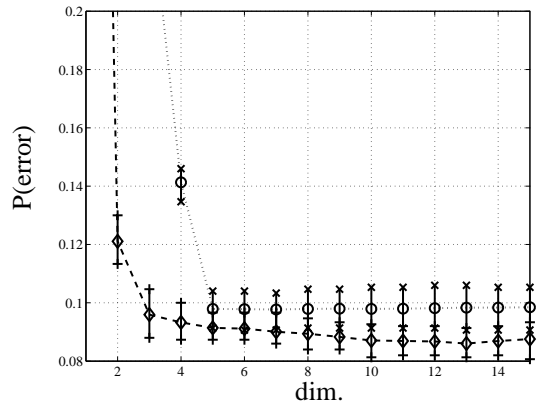}
 }
\caption{Probability of incorrect classification vs. CCDR's
 dimension for a linear classifier (dotted line $\circ$) and
 for the $k$-nearest neighbors algorithm (dashed line $\diamond$) preprocessed by CCDR.
 $80\%$ confidence intervals are presented as $\times$ for the linear classifier and as $+$
 for the $k$-nearest neighbors algorithm.}
 \label{fig:vs:dim}
\end{figure}
While the data points in $\cX_n$ may lie on a manifold of a
particular dimension, the actual dimension required for
classification may be smaller. Here, we examine classification
performance as a function of the CCDR dimension. Using the
entropic graph  dimension estimation algorithm in
\cite{costa:04:tsp}, we obtain the following  estimated dimension
for each class:
\begin{center}
\begin{tabular}{|c ||c |c |c |c |c |c|} \hline
class & 1 & 2 & 3 & 4 & 5 & 6 \\ \hline \hline dimension & 13 &  7
& 13 & 10 &6 & 13 \\ \hline
\end{tabular}

\end{center}
Therefore, if an optimal nonlinear embedding of the data could be
found, we suspect that a dimension greater than $13$ may not yield
significant improvement in classification performance. Since CCDR
does not necessarily yield an optimal embedding, we choose CCDR
embedding dimension as $d=14$ in Section \ref{sec:beta}.

In Fig.~\ref{fig:vs:dim}, we plot the classification error
probability (dotted line) vs. CCDR dimension and its confidence
interval for a linear classifier. We observed decrease in error
probability as the dimension increases. When the CCDR dimension is
greater than $5$, the error probability seems fairly constant.
This is an indication that CCDR dimension of $5$ is sufficient for
classification if one uses the linear classifier with $\beta=0.5$,
i.e., linear classifier cannot exploit geometry.

We also plot the classification error probability (dashed line)
vs. CCDR dimension and its confidence interval for $k$-nearest
neighbors classifier. Generally, we observe decrease in error
probability as the dimension increases. When the CCDR dimension is
greater than $5$, the error probability seems fairly constant.
When CCDR dimension is three, classifier error is below $0.1$. On
the other hand, minimum possibility of error obtained at CCDR
dimension 12-14. This is remarkable agreement with the dimension
estimate of $13$ obtained using the entropic graph algorithm of
\cite{costa:04:tsp}.

\subsection{CCDR's $k$-Nearest Neighbors Parameter}
\begin{figure}[htb!]
 \centering
 \resizebox{!}{0.33\textwidth}{
 \psfrag{K-CCDR}{K-CCDR}
 \psfrag{P(error)}{P(error)}
 \includegraphics[width=0.5\textwidth]{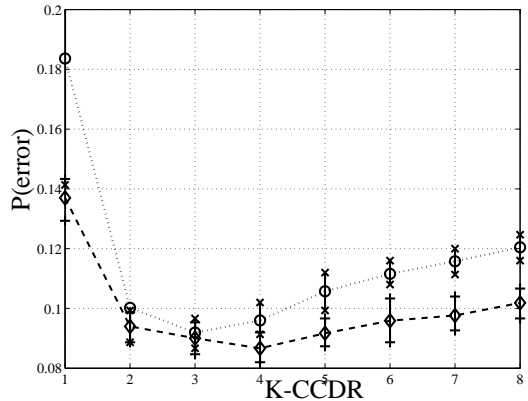}
 }
\caption{Probability of incorrect classification vs. CCDR's
 $k$-nearest neighbors parameter for a linear classifier (dotted line $\circ$) and
 for the $k$-nearest neighbors algorithm (dashed line $\diamond$) preprocessed by CCDR.
 $80\%$ confidence intervals are presented as $\times$ for the linear classifier and as $+$
 for the $k$-nearest neighbors algorithm.}
 \label{fig:vs:k}
\end{figure}
The last parameter we examine is the CCDR's $k$-nearest neighbors
parameter. In general, as $k$ increases non-local distances are
included in the lower-dimensional embedding. Hence, very large $k$
prevents the flexibility necessary for dimensionality reduction on
(globally) non-linear (but locally linear) manifolds.

In Fig.~\ref{fig:vs:k}, the classification probability of error
for the linear classifier (dotted line) is plotted vs. the CCDR's
$k$-nearest neighbors parameter. A minimum is obtained at $k=3$
with probability of error of $0.092$. The classification
probability of error for $k$-nearest neighbors (dashed line) is
plotted vs. the CCDR's $k$-nearest neighbors parameter. A minimum
is obtained at $k=4$ with probability of error of $0.086$.


\section{Conclusion}
\label{sec:conclusion}

In this paper, we presented the CCDR algorithm for multiple
classes. We examined the performance of various classification
algorithms applied after CCDR for the Landsat MSS imagery dataset.
We showed that for a linear classifier, decreasing $\beta$ yields
improved performance and for a $k$-nearest neighbors classifier,
increasing $\beta$ yields improved performance. We demonstrated
that both classifiers have improved performance on the much
smaller dimension of CCDR embedding space than when applied to the
original high-dimensional data. We also explored the effect of $k$
in the $k$-nearest neighbors construction of CCDR weight matrix on
classification performance.  CCDR allows reduced complexity
classification  such as the linear classifier to perform better
than more complex classifiers applied to the original data. We are
currently pursuing an out-of-sample extension to the algorithm
that does not require rerunning CCDR on test and training data to
classify new test point.

\bibliographystyle{IEEEtran}
\bibliography{ccdr}

\begin{thebibliography}{10}
\providecommand{\url}[1]{#1}
\csname url@rmstyle\endcsname
\providecommand{\newblock}{\relax}
\providecommand{\bibinfo}[2]{#2}
\providecommand\BIBentrySTDinterwordspacing{\spaceskip=0pt\relax}
\providecommand\BIBentryALTinterwordstretchfactor{4}
\providecommand\BIBentryALTinterwordspacing{\spaceskip=\fontdimen2\font plus
\BIBentryALTinterwordstretchfactor\fontdimen3\font minus
  \fontdimen4\font\relax}
\providecommand\BIBforeignlanguage[2]{{%
\expandafter\ifx\csname l@#1\endcsname\relax
\typeout{** WARNING: IEEEtran.bst: No hyphenation pattern has been}%
\typeout{** loaded for the language `#1'. Using the pattern for}%
\typeout{** the default language instead.}%
\else
\language=\csname l@#1\endcsname
\fi
#2}}

\bibitem{hastie:00}
T.~Hastie, R.~Tibshirani, and J.~Friedman, \emph{The Elements of Statistical
  Learning Data Mining, Inference, and Prediction}, ser. Springer Series in
  Statistics.\hskip 1em plus 0.5em minus 0.4em\relax New York: Springer Verlag,
  2000.

\bibitem{Jain&Dubes:88}
A.~K. Jain and R.~C. Dubes, \emph{Algorithms for clustering data}.\hskip 1em
  plus 0.5em minus 0.4em\relax New Jersey: Prentice Hall, 1998.

\bibitem{Torgerson:52}
W.~S. Torgerson, ``Multidimensional scaling: I. theory and method,''
  \emph{Psychometrika}, vol.~17, pp. 401--419, 1952.

\bibitem{Cox&Cox:00}
T.~F. Cox and M.~A.~A. Cox, \emph{Multidimensional Scaling}, 2nd~ed., ser.
  Monographs on Statistics and Applied Probability.\hskip 1em plus 0.5em minus
  0.4em\relax London: Chapman \& Hall/CRC, 2000, vol.~88.

\bibitem{Scholkopf:98:nc}
B.~Sch{\"o}lkopf, A.~J. Smola, and K.-R. M{\"u}ller, ``Nonlinear component
  analysis as a kernel eigenvalue problem.'' \emph{Neural Computation},
  vol.~10, no.~5, pp. 1299--1319, 1998.

\bibitem{aizerman:64:arc}
A.~Aizerman, E.~M. Braverman, and L.~I. Rozoner, ``Theoretical foundations of
  the potential function method in pattern recognition learning,''
  \emph{Automation and Remote Control}, vol.~25, pp. 821--837, 1964.

\bibitem{Tenenbaum:00:science}
J.~B. Tenenbaum, V.~D. Silva, and J.~C. Langford, ``A global geometric
  framework for nonlinear dimensionality reduction.'' \emph{Science}, vol. 290,
  no. 5500, pp. 2319--2323, 2000.

\bibitem{Belkin&Niyogi:NC03}
M.~Belkin and P.~Niyogi, ``Laplacian eigenmaps for dimensionality reduction and
  data representation,'' \emph{Neural Computation}, vol.~15, no.~6, pp.
  1373--1396, June 2003.

\bibitem{Roweis:00:science}
S.~T. Roweis and L.~K. Saul, ``Nonlinear dimensionality reduction by locally
  linear embedding,'' \emph{Science}, vol. 290, no. 5500, pp. 2323--2326,
  December 2000.

\bibitem{Donoho:03:PNAS}
D.~L. Donoho and C.~Grimes, ``{H}essian eigenmaps: {L}ocally linear embedding
  techniques for high-dimensional data,'' \emph{Proceedings of the National
  Academy of Sciences of the United States of America}, vol. 100, no.~10, pp.
  5591--5596, May 2003.

\bibitem{zhang:05:siam}
Z.~Zhang and H.~Zha, ``Principal manifolds and nonlinear dimensionality
  reduction via tangent space alignment,'' \emph{SIAM J. Sci. Comput.},
  vol.~26, no.~1, pp. 313--338, 2005.

\bibitem{coifman:06:ACHA}
R.~Coifman and S.~Lafon, ``Diffusion maps,'' \emph{Applied and Computational
  Harmonic Analysis: Special issue on Diffusion Maps and Wavelets}, vol.~21,
  pp. 5--30, July 2006.

\bibitem{weinberger:04:cvpr}
K.~Q. Weinberger and L.~K. Saul, ``Unsupervised learning of image manifolds by
  semidefinite programming,'' in \emph{Proceedings of the IEEE Conference on
  Computer Vision and Pattern Recognition (CVPR-04)}, vol.~2, Washington D.C.,
  2004, pp. 988--995.

\bibitem{costa:05:icassp}
J.~A. Costa and A.~O.~H. III, ``Classification constrained dimensionality
  reduction,'' in \emph{Proc. IEEE Intl. Conf. on Acoust., Speech, and Signal
  Processing}, vol.~5, March 2005, pp. 1077--1080.

\bibitem{yan:07:pami}
S.~Yan, D.~Xu, B.~Zhang, H.-J. Zhang, Q.~Yang, and S.~Lin, ``Graph embedding
  and extensions: A general framework for dimensionality reduction,''
  \emph{IEEE Trans. on Pattern Analysis and Machine Intelligence}, vol.~29,
  no.~1, pp. 40--51, Jan. 2007.

\bibitem{raich:06:icassp}
R.~Raich, J.~A. Costa, and {A.~O.~Hero~III}, ``On dimensionality reduction for
  classification and its application,'' in \emph{Proc. IEEE Intl. Conf.
  Acoust., Speech, Signal Processing}, vol.~5, Toulouse, France, May 2006, pp.
  917--920.

\bibitem{Bengio:03}
Y.~Bengio, J.-F. Paiement, and P.~Vincent, ``Out-of-sample extensions for
  {LLE}, isomap, {MDS}, eigenmaps, and spectral clustering,'' D\'{e}partement
  d'Informatique et Recherche Op\'{e}rationnelle Universit\'{e} de Montr\'{e}al
  Montr\'{e}al, Qu\'{e}bec, Canada, H3C 3J7, Tech. Rep., 2003, technical Report
  1238, D\'{e}partement d'Informatique et Recherche Op\'{e}rationnelle.

\bibitem{Bengio:04}
Y.~Bengio, O.~Delalleau, N.~Le~Roux, J.~F. Paiement, P.~Vincent, and M.~Ouimet,
  ``Learning eigenfunctions links spectral embedding and kernel {PCA},''
  \emph{Neural Computation}, vol.~16, no.~10, pp. 2197--2219, 2004.

\bibitem{sat:data}
``Satellite image data,'' available at {\footnotesize
  \texttt{http://www.liacc.up.pt/ML/statlog/datasets/
  satimage/satimage.doc.html}}.

\bibitem{costa:04:tsp}
J.~A. Costa and A.~O. Hero, ``Geodesic entropic graphs for dimension and
  entropy estimation in manifold learning,'' \emph{IEEE Trans. Signal
  Processing}, vol.~52, no.~8, pp. 2210--2221, Aug. 2004.

\end{thebibliography}





\end{document}